**Operational digital twin clinics enable task-based evaluation of embodied AI**

**Authors**: Xinyuan Wu, MD[1], Jingrao Zhang, MS[2], Mengdi Xu, PhD[3], Henry K. Chu, PhD[4], Mingguang He, MD, PhD[1,5,6]*, and Danli Shi, MD, PhD[1,5]*.

**Affiliations**:

1. School of Optometry, The Hong Kong Polytechnic University, Hong Kong SAR, China.
2. Faculty of Science and Technology, University of Macau, Macau, China.
3. Institute for Interdisciplinary Information Sciences, Tsinghua University, Beijing, China.
4. Department of Mechanical Engineering, The Hong Kong Polytechnic University, Hong Kong SAR, China.
5. Research Centre for SHARP Vision (RCSV), The Hong Kong Polytechnic University, Hong Kong SAR, China.
6. Centre for Eye and Vision Research (CEVR), 17W Hong Kong Science Park, Hong Kong SAR, China.

**Correspondence:**

Dr. Danli Shi, The Hong Kong Polytechnic University, Hong Kong, China. Email: danli.shi@polyu.edu.hk

Prof. Mingguang He, Chair Professor of Experimental Ophthalmology, The Hong Kong Polytechnic University, Hong Kong, China. Email: mingguang.he@polyu.edu.hk

**Abstract**

Embodied artificial intelligence (AI) must be tested in the clinical environments where it will operate, but building realistic, robot-testable settings is costly and difficult to scale. Here we show that routine clinic images can be transformed into operational digital twins for task-based evaluation of embodied AI. Using 39 ophthalmic clinic scenes, we converted single photographs into editable, simulator-ready environments and assessed reconstruction quality, room-scale geometry, mesh grounding, multi-robot feasibility, perturbation sensitivity and closed-loop policy performance. The reconstructed scenes preserved workspace structure, while local editing enabled controlled device reconfiguration. Device meshes, collision proxies and semantic anchors converted visual reconstructions into contact-aware simulation scenes. Across three robot embodiments, shared task targets showed different patterns of reachability and contact feasibility. Small device translations and rotations produced task-specific changes in contact margins that were not captured by visual similarity alone. Digital-twin trajectories also supported local policy learning and closed-loop evaluation. These findings establish operational validity as a key principle for clinical digital twins and provide an intermediate layer between offline development and physical deployment of embodied AI in healthcare.

**Key Words**

Operational digital twin; ophthalmic robotics; embodied AI; 3D Gaussian Splatting; contact-aware simulation.

## Introduction

Embodied artificial intelligence (EAI) extends medical automation from interpretation to action in physical space.[1] Diagnostic models can be evaluated on images or clinical records, but robots must navigate real environments, approach people and equipment, and interact with objects whose position and surroundings vary across clinics.[2-6] This challenge is especially relevant in outpatient care, where tasks are performed in shared, equipment-dense rooms rather than purpose-built robotic workcells.[7,8] Before physical testing, robots therefore require simulation environments that preserve the clinic features relevant to action, including room layout, device placement, patient support, approach paths and task-specific contact surfaces.

Current clinical simulators do not fully meet this need. Most are designed for human training, procedural rehearsal or visual demonstration rather than robot learning and evaluation.[9] They often lack the scene-level geometry needed for localization, motion planning, contact reasoning and policy assessment.[10,11] This gap is particularly important in outpatient settings, where equipment positions, patient placement and staff movement vary during routine care.[3,4] In contrast, most medical EAI simulation research has focused on surgery and other highly controlled settings, limiting its relevance to outpatient automation.[12,13]

Ophthalmic clinics provide a demanding testbed. Slit lamps, tonometers, retinal imaging devices, chairs, tables and displays are often arranged in small rooms with narrow approach paths and workflow-dependent layouts.[14-16] Small changes in device position can alter robot reachability, collision risk and contact margins. These characteristics make ophthalmic clinics well suited for studying how embodied systems can be evaluated before deployment near patients and clinical equipment.

Digital twins offer a potential solution,[17,18] but healthcare digital-twin research has largely focused on physiology or health-system modelling rather than physical clinical environments. Building site-specific environments remains difficult to scale.[19-21] Manual modelling is time-consuming and requires specialist expertise, whereas multi-view scanning can interrupt clinical workflow and raise privacy concerns.[22,23] A practical outpatient digital twin should be inexpensive to create, preserve site-specific context, allow local reconfiguration and provide geometry that robots can query.

Recent generative 3D methods offer a low-burden route to reconstructing clinical scenes from limited observations.[24-27] However, visual realism alone is insufficient. A scene may appear convincing but still lack reliable collision geometry, task anchors or contact margins. Conversely, an imperfect reconstruction may be operationally useful if it preserves the spatial relationships, support surfaces and device geometry needed for a defined task. The key question is therefore not whether an image-derived scene is visually perfect, but whether it can support meaningful robot-facing evaluation.

Here we developed a workflow for constructing operational digital twins of ophthalmic clinics. Starting from routine clinic images, the workflow generates neural scene representations, converts them into simulator environments and evaluates them using clinician-supervised task definitions. Across 39 scenes, we assessed reconstruction and simulator conversion, local scene editing, mesh-grounded contact geometry, multi-robot task feasibility, sensitivity to instrument perturbations and scene-specific policy evaluation. Using ophthalmic outpatient clinics as a demanding testbed, we show how routine images can become editable simulation environments for contact-proxy testing, perturbation analysis and staged evaluation before physical deployment.

## Results

This study evaluated a staged workflow for constructing operational digital twins from routine ophthalmic clinic images (**Fig. 1**). We examined whether single images could produce simulator-ready scenes, support local reconfiguration, enable contact-aware task definition, reveal embodiment-specific feasibility and sensitivity to device placement, and provide trajectories for policy evaluation. **Supplementary Video 1** demonstrates the workflow, including reconstruction, editing, mesh grounding, inverse-kinematics (IK) probing, perturbation testing and closed-loop rollout.

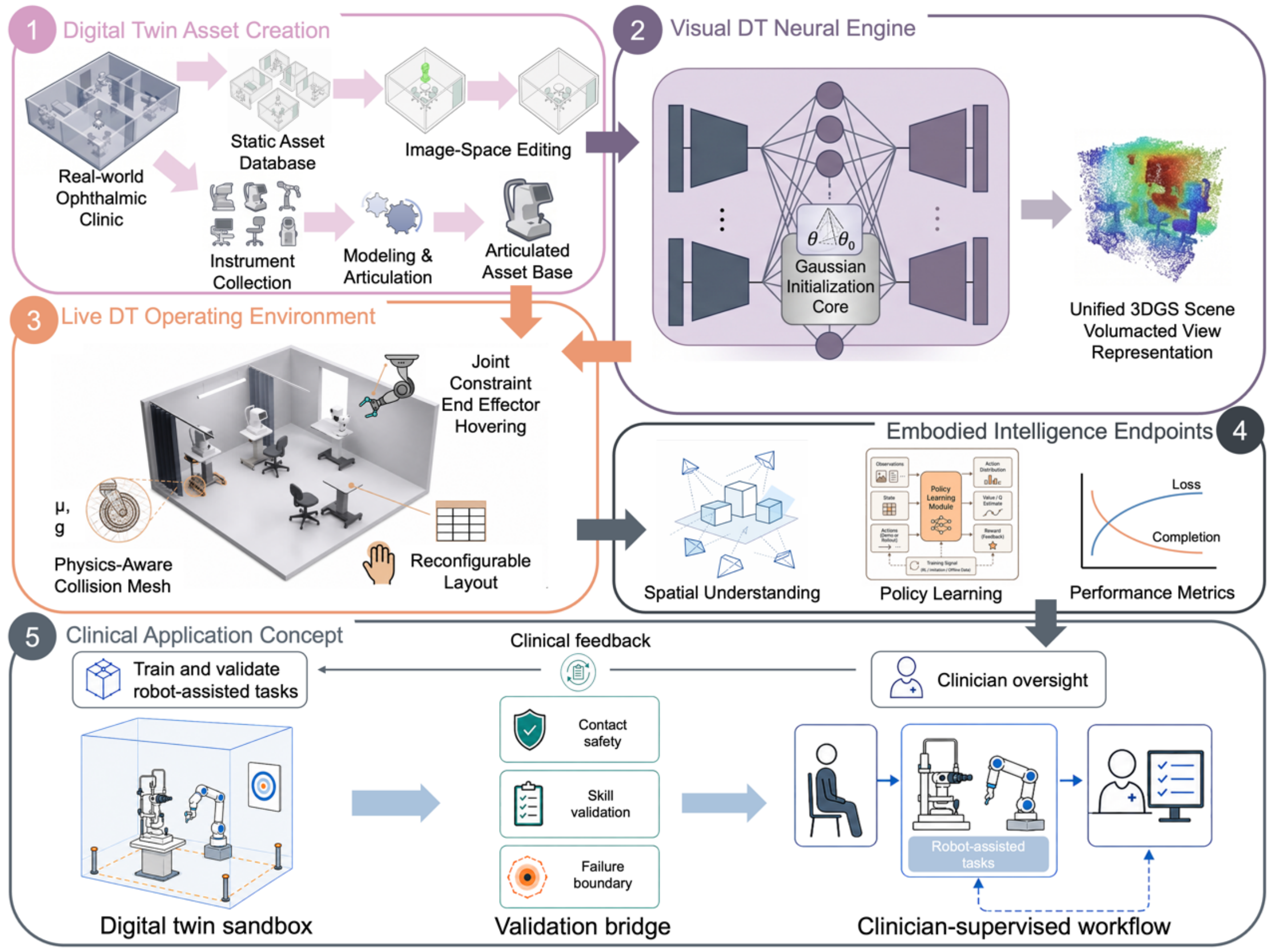

**Figure 1. From routine clinic images to operational digital twins for embodied AI evaluation.** Routine ophthalmic clinic images are converted into operational digital twins through five stages. First, real-world clinic images are collected and used to define the clinical workspace; target ophthalmic instruments are prepared as digital device assets, with optional image-space editing for local reconfiguration. Second, single-image 3D Gaussian splatting (3DGS) generates a neural scene representation. Third, reconstructed scenes are instantiated in a physics-aware simulator with device meshes, collision geometry, articulated components and task anchors. Fourth, the digital twin supports robot-facing evaluation, including spatial analysis, inverse-kinematics (IK) contact probing, policy learning and performance assessment. Finally, the workflow provides a sandbox for task-based evaluation of robot-assisted clinical interactions, including contact safety, task feasibility and failure boundaries.

### Image-derived clinic reconstruction

We reconstructed 39 routine clinic photographs, including examination rooms, waiting areas and corridor-like spaces, using single-image 3D Gaussian splatting (3DGS), detailed in **Supplementary Table 1**. Reconstructions preserved major workspace features, including room layout, equipment silhouettes, tabletop devices and coarse depth ordering (**Fig. 2A, Extended Data Fig. 1**).

Observed-view reconstruction quality was high across scenes, with mean SSIM of 0.908 ± 0.005, LPIPS of 0.034 ± 0.001, CLIP similarity of 0.969 ± 0.006 and PSNR of 31.99 ± 0.30 (Fig. 2B). Supportive measures showed FID of 60.52 and DISTS of 0.0168 ± 0.0013 (**Supplementary Table 2**). Circular render trajectories showed generally stable cross-view structure, with larger variation mainly in highly occluded views (**Extended Data Fig. 2**).

Reconstructions also retained room-scale geometry relevant to robot interaction. Ground-plane fitting yielded a plane RMSE of 8.54 ± 0.24 mm, and dominant-surface normal error was 13.77 ± 1.23°. LiDAR-referenced depth agreement showed an AbsRel of 0.0869 ± 0.0063, RMSE of 0.396 ± 0.034 m, MAE of 0.251 ± 0.026 m, Log10 error of 0.0366 ± 0.0027 and $\delta < 1.25$ of 0.930 ± 0.010 (Fig. 2C,D). Depth residuals were concentrated near thin structures and occlusion boundaries rather than across the main room envelope (**Extended Data Fig. 3**).

Simulator conversion retained the dominant scene structure required for downstream scene assembly (**Fig. 2E** and **Extended Data Fig. 4** and **Supplementary Fig. 1**). Mean symmetric Chamfer similarity was 0.773 ± 0.013, cyclic alignment similarity was 0.718 ± 0.013 and structural preservation was 0.878 ± 0.006 (**Supplementary Table 3**). Using prespecified descriptive cutoffs, 74.4% of scenes met the visual-similarity criterion, 69.2% met the cyclic-alignment criterion, 87.2% met the structural-preservation criterion and 61.5% met all three criteria (**Fig. 2F** and **Supplementary Table 4**). These results indicate that routine clinic images can preserve sufficient spatial and device context for simulator instantiation and robot-facing evaluation.

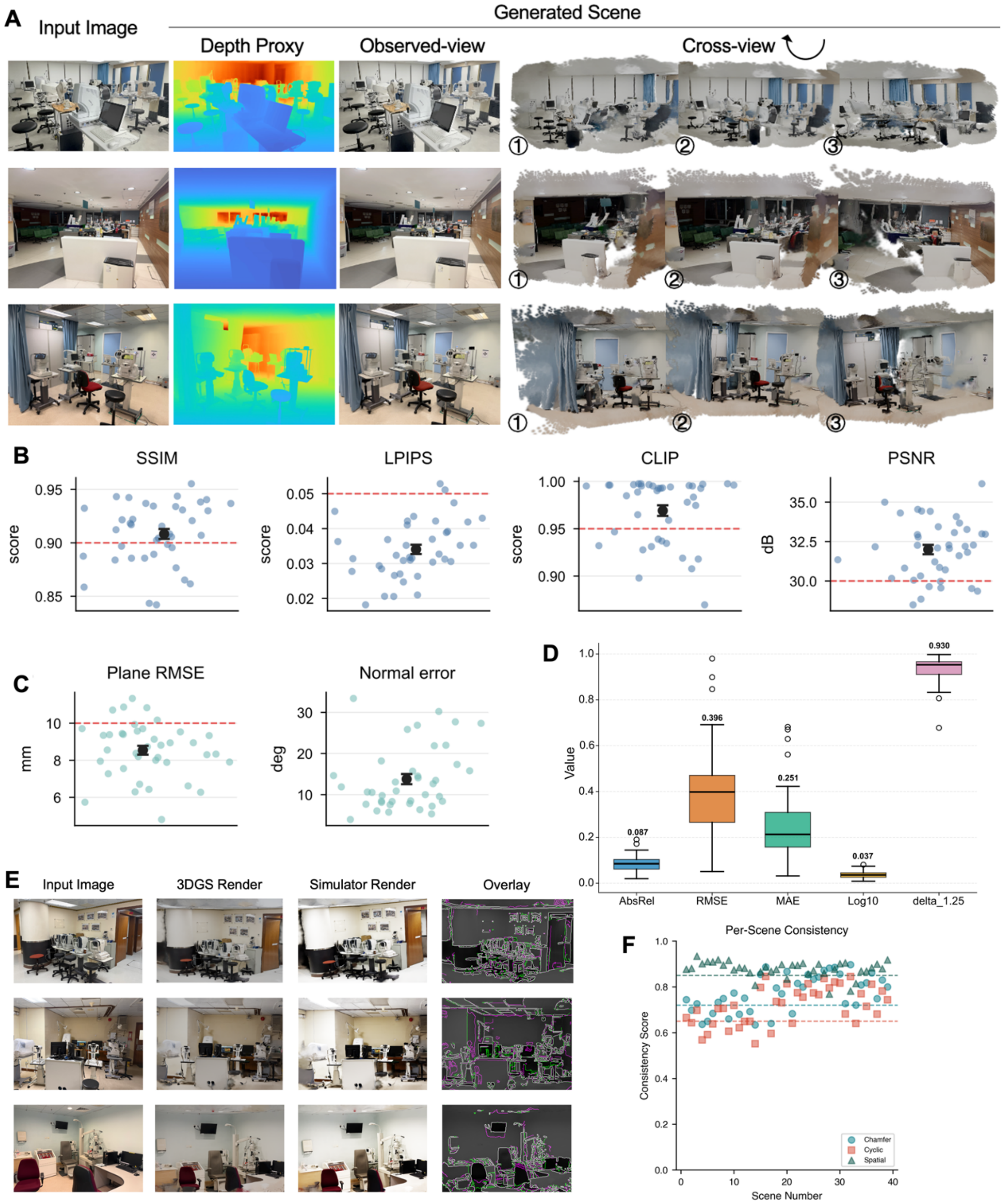

**Figure 2. Routine images preserve clinic structure for simulator instantiation.** (A) Reconstruction workflow from input image to depth proxy, observed-view reconstruction and cross-view rendering. (B) Observed-view quality across 39 scenes measured by SSIM, LPIPS, CLIP similarity and PSNR. (C) Room-scale geometry assessed by ground-plane RMSE and dominant-surface normal error. (D) LiDAR-referenced depth agreement measured by AbsRel, RMSE, MAE, Log10 error and δ<1.25. (E) Matched renderings and registered edge overlays; white contours indicate agreement and coloured contours indicate local mismatch. (F) Per-scene conversion scores for symmetric Chamfer similarity, cyclic alignment similarity and structural preservation (1 − smoothness gap). Dashed lines indicate prespecified cutoffs.

**Editable clinic configurations**

We next tested whether scenes could be locally reconfigured before reconstruction. In 32 scenes containing a visible clinical device, image-space editing removed or replaced the target region while preserving the surrounding room and non-target equipment (**Fig. 3A**).

Unmasked-region SSIM was 0.977, whereas masked-region SSIM was 0.526, indicating that editing was concentrated within the intended target region (**Fig. 3B)**. Global and local FID were 79.9 and 267.9, respectively, consistent with greater distributional change in the edited area (**Fig. 3C**). After reconstruction, the absolute shift in scene-quality score between raw and edited scenes was 0.073 ± 0.016, and 84.4% of edited scenes remained within the prespecified preservation boundary of 0.15 (**Fig. 3D** and **Supplementary Table 5**). Depth renderings remained visually stable outside the edited region (**Extended Data Fig. 5**). These results show that local device changes can be introduced without destabilizing the broader clinic context, enabling controlled reconfiguration of site-specific digital twins.

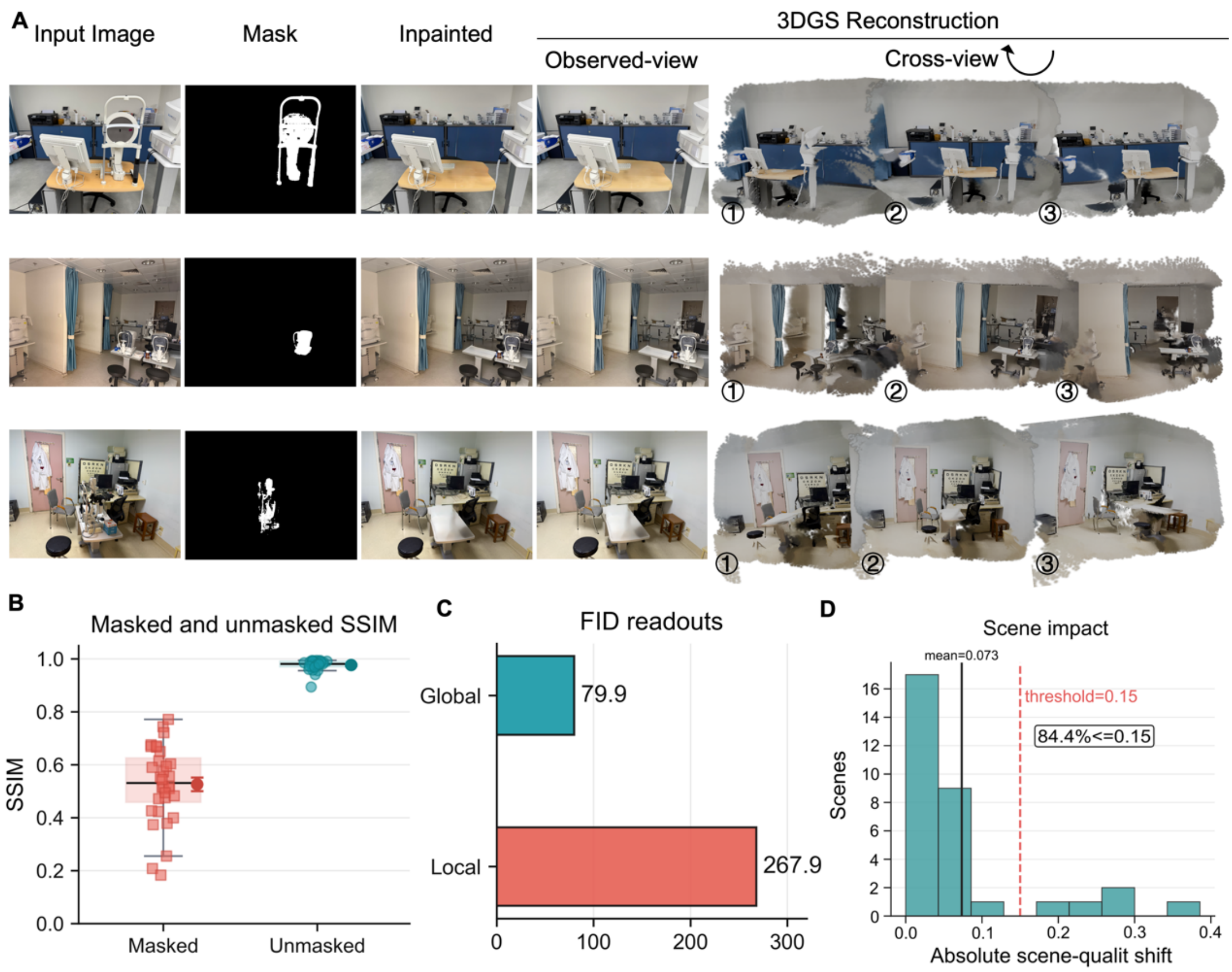

**Figure 3. Local editing enables controlled clinical reconfiguration without disrupting the surrounding workspace.** (A) Representative editing workflow

showing the input image, editing mask, inpainted image, observed-view reconstruction and cross-view renderings after edited-scene 3DGS reconstruction. (B) Masked- and unmasked-region SSIM for the semi-automated editing workflow. Bars show means, error bars show s.e.m. and points represent individual scenes. (C) Global and local FID; local FID was calculated from crops centred on the edited region. (D) Distribution of the absolute change in scene-quality score between raw and edited reconstructions. The dashed line indicates the prespecified preservation boundary (0.15); the solid line indicates the cohort mean.

**Mesh-grounded contact geometry**

Visual reconstruction alone does not provide the geometry needed for robot interaction. We therefore integrated clinical-device meshes, conservative collision proxies and semantic task anchors into selected reconstructed rooms (**Fig. 4A** and **Extended Data Fig. 6A**). The reconstructed scene retained site-specific spatial context, whereas the mesh layer supplied task-relevant device geometry for screen and handle contact-proxy tasks.

Across seven mesh-grounded scenes, visible collider occupancy averaged $4.76 \pm 0.36\%$ and remained below 7% in every scene (**Fig. 4B**). This allowed collision geometry to be incorporated without obscuring the reconstructed environment.

We assessed the resulting interaction geometry using 58 filtered reachability traces, of which 52 formed the dominant cluster. The contact surface showed an in-plane spread of 47.1 mm, normal-direction spread of 5.7 mm and plane-fit RMSE of 9.6 mm; the normal-to-in-plane dispersion ratio was 12.2% (**Fig. 4C,D** and **Extended Data Fig. 6B,C**). Thus, mesh grounding converted image-derived scenes into operational environments that could support access and contact assessment.

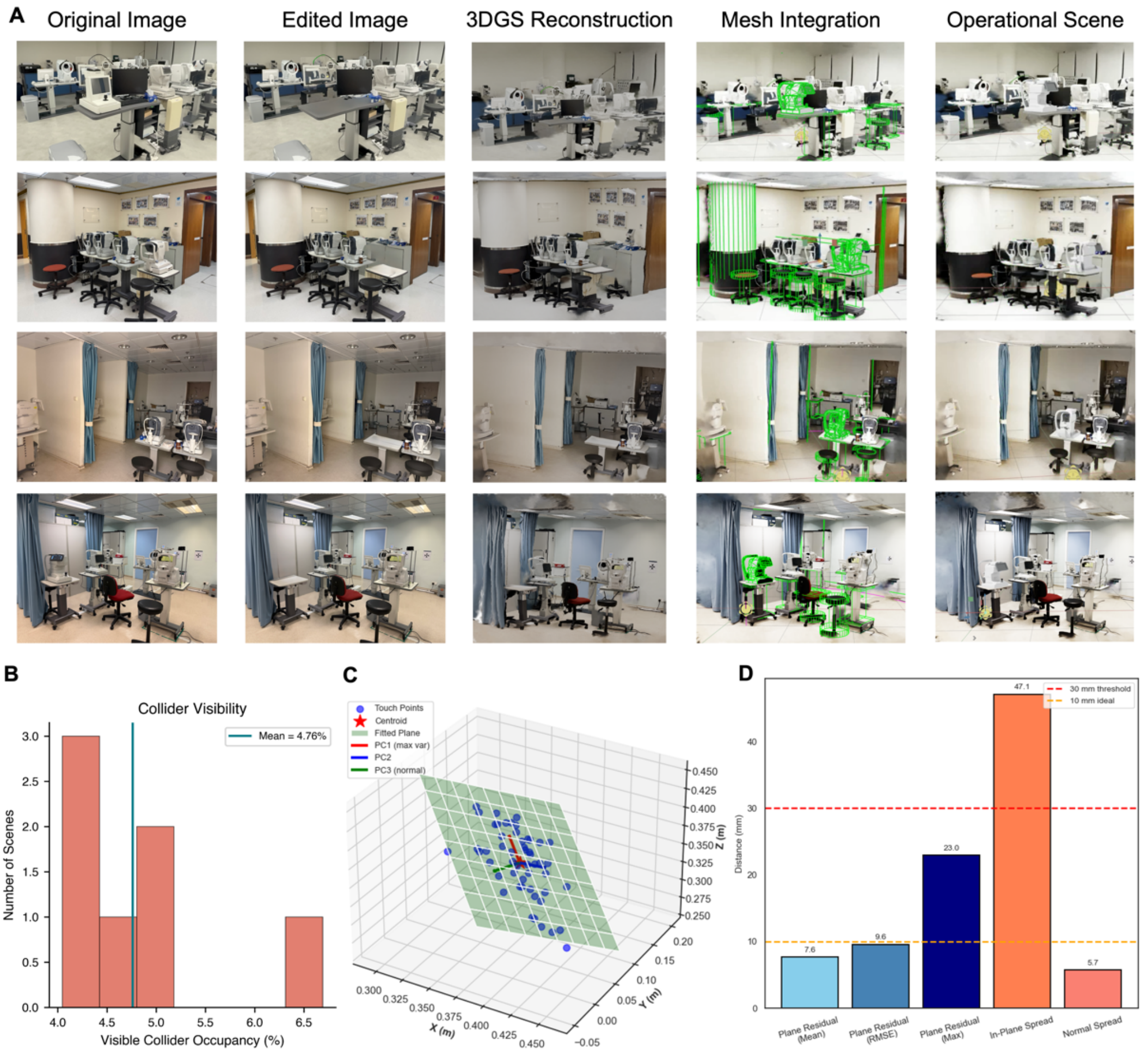

**Figure 4. Mesh grounding converts visual reconstructions into contact-aware operational scenes.** (A) Reconstruction-derived clinic context is combined with device meshes, collision geometry and task anchors to create an operational simulation environment. (B) Visible collider occupancy across seven mesh-grounded scenes; the vertical line indicates the cohort mean. (C) Principal-component analysis of the dominant interaction cluster from simulator reachability trajectories, showing contact points, fitted plane and principal axes. (D) Plane-fit residuals and directional spread of the dominant interaction cluster. Dashed lines indicate heuristic guide values for operational scene assessment. This step links image-derived scene context to robot-queryable geometry for contact-proxy evaluation.

## Embodiment-specific feasibility

We used IK trajectory collection to test whether the same task anchors could be queried across three robot embodiments: SO101, Kinova Gen3 and Franka Panda. The benchmark included three matched operational scene variants and two contact-proxy tasks: handle contact and screen contact (**Fig. 5A** and **Supplementary Fig. 2**).

SO101 achieved feasibility rates of 99.3% for handle contact and 98.3% for screen contact. Kinova Gen3 achieved 79.0% and 94.2%, respectively, whereas Franka Panda achieved 85.7% and 93.7% (**Fig. 5B**). Feasibility varied across scene–robot combinations: Gen3 handle feasibility fell to 58.0% in scene 1, Gen3 screen feasibility was 84.0% in scene 2 and Franka handle feasibility was lowest in scene 3 at 75.0% (**Fig. 5C**).

Contact diagnostics further showed scene- and embodiment-dependent margins. Handle contact-distance distributions varied across robots and scenes, while screen residuals showed differences in plane-normal contact accuracy (**Fig. 5D,E**). Among failed trials, the most common causes were mesh not reached (59.2%), mesh out of reach (21.5%) and safety-gated penetration (15.1%) (**Extended Data Fig. 7**). These findings show that the same clinical target can be feasible or infeasible for different geometric reasons depending on robot embodiment, base placement, end-effector geometry, reach envelope and collision constraints.

We next asked whether small changes in device pose altered these feasibility margins. In one representative scene, the instrument mesh was translated by ±20 mm along local x and y axes or rotated by ±5° in yaw, with task anchors recomputed after each perturbation (**Fig. 5F** and **Extended Data Fig. 8A**). Screen contact was more configuration sensitive than handle contact. SO101 screen success ranged from 30% to 100%, Kinova Gen3 from 60% to 100%, and Franka Panda dropped from 100% in most conditions to 0% under +20 mm local-y translation (**Fig. 5G,H**). By contrast, handle-contact success remained comparatively stable, staying at or above 75% across all tested perturbations (**Fig. 5G,I**). These results show that operational digital twins can expose task- and embodiment-specific reachability and contact-margin boundaries before physical robot testing.

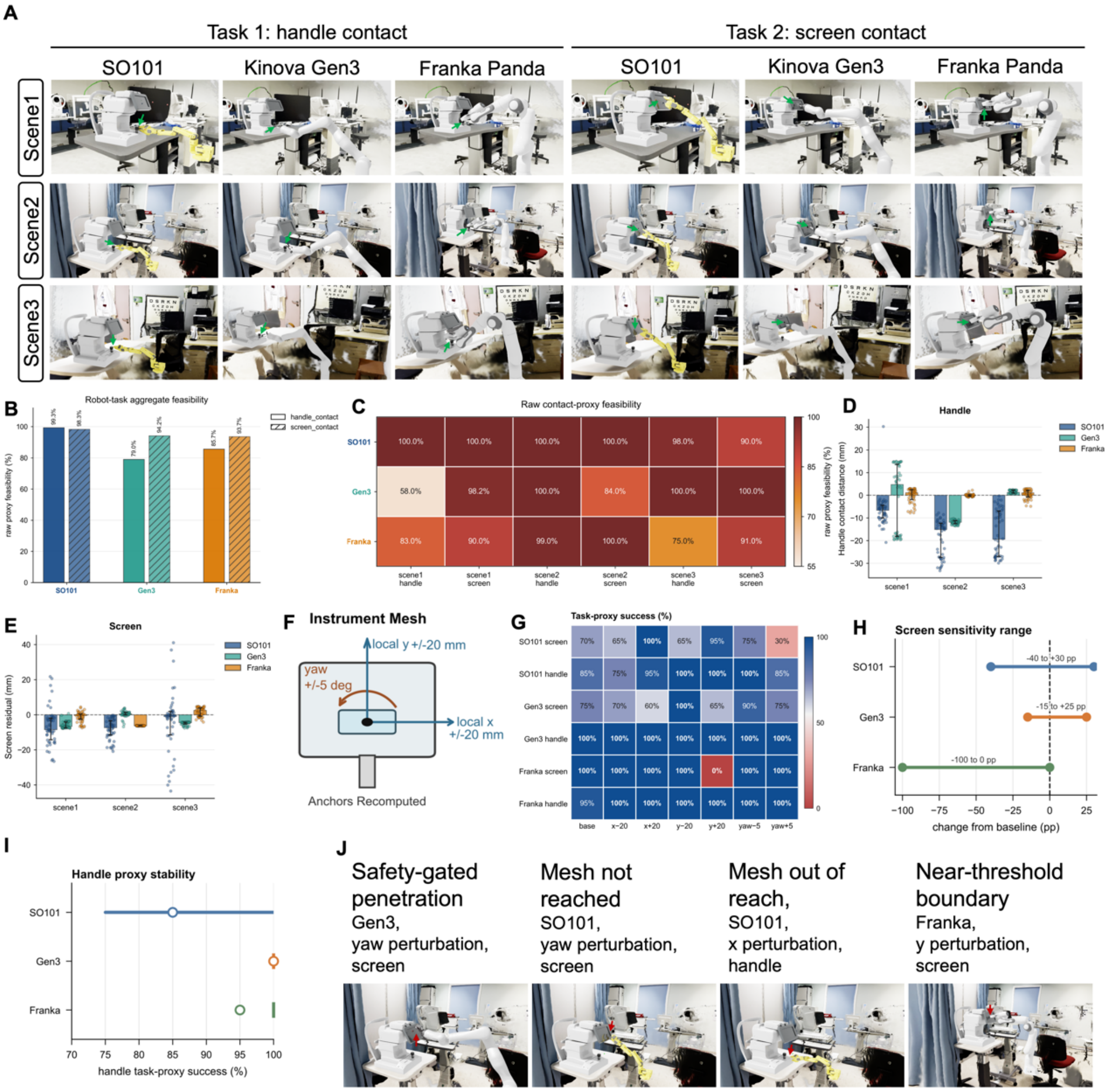

**Figure 5. Shared clinical targets reveal embodiment-specific feasibility and contact-margin boundaries.** (A) Representative handle- and screen-contact proxy trajectories for SO101, Kinova Gen3, and Franka Panda across matched operational scenes. Green arrows indicate successful contact sites. (B) Contact-proxy feasibility by robot and task. (C) Robot-task-scene heat map showing cell-level success rates. (D, E) Contact-quality diagnostics for selected trajectories, including handle-contact distance and screen plane-residual measures. (F) Perturbation protocol. Instrument meshes were translated by ±20 mm along local x and y axes or rotated by ±5° in yaw; screen and handle anchors were recomputed for each condition. (G) Task-proxy success by robot, task and perturbation condition. (H) Percentage-point change in success relative to baseline. (I) Handle-contact success across perturbation conditions. (J) Representative boundary cases, including safety-gated penetration, target not reached, out-of-reach targets and near-threshold contact. Red arrows indicate the end-effector position in failed or boundary cases.

**Policy learning and evaluation**

Finally, we tested whether trajectories collected in operational digital twins could support local policy learning. Using the SO101 scene, we trained joint-only behaviour cloning, target-conditioned behaviour cloning, conditional diffusion policy and residual behaviour cloning for screen-touch and handle-contact tasks (**Fig. 6A**).

The handle dataset included 104 training trajectories and 26 validation trajectories, comprising 4,000 and 1,062 frames, respectively. The screen dataset included 2,145 training trajectories and 509 validation trajectories, comprising 35,125 and 7,274 frames (**Extended Data Fig. 9A,B**). Target-conditioned behaviour cloning improved one-step action MAE relative to joint-only behaviour cloning for both screen contact (1.92 versus 2.85 mm) and handle contact (1.58 versus 1.80 mm). Diffusion policy achieved MAEs of 1.96 mm for screen contact and 2.10 mm for handle contact (**Fig. 6B**).

During offline rollout, residual behaviour cloning achieved the lowest mean final error for screen contact (33.3 mm), compared with 66.4 mm for target-conditioned behaviour cloning and 64.0 mm for diffusion policy. For handle contact, residual behaviour cloning achieved a mean final error of 3.7 mm, whereas target-conditioned behaviour cloning was unstable at the plotted scale (**Fig. 6C,D**).

Closed-loop simulator rollout added runtime scene geometry, simulator stepping, finite-difference IK execution and task-specific success gates. Screen-touch success reached 73.1%, and handle-contact success reached 69.2% (**Fig. 6E**). Error decreased over the course of rollout, although failures remained associated with target non-reachability and incorrect screen-contact locations (**Fig. 6F,G** and **Supplementary Figs. 3 and 4**). Together, these results show that operational digital twins can provide task-specific trajectories for policy learning and closed-loop predeployment evaluation.

Three handle and three screen trajectories were each replayed five times on a non-patient SO101 platform, with paired videos showing physical execution in a clinic-like arrangement and commanded joint profiles shown in **Extended Data Fig.10A-C**. The handle-contact policy was further evaluated in five online executions, where runtime logs captured target-directed kinematic progress and paired videos documented visible contact in selected runs (**Extended Data Fig. 10D,E**). Together, these hardware checks extended the digital-twin clinic workflow from simulation-based trajectory and policy evaluation to an initial stage of physical execution.

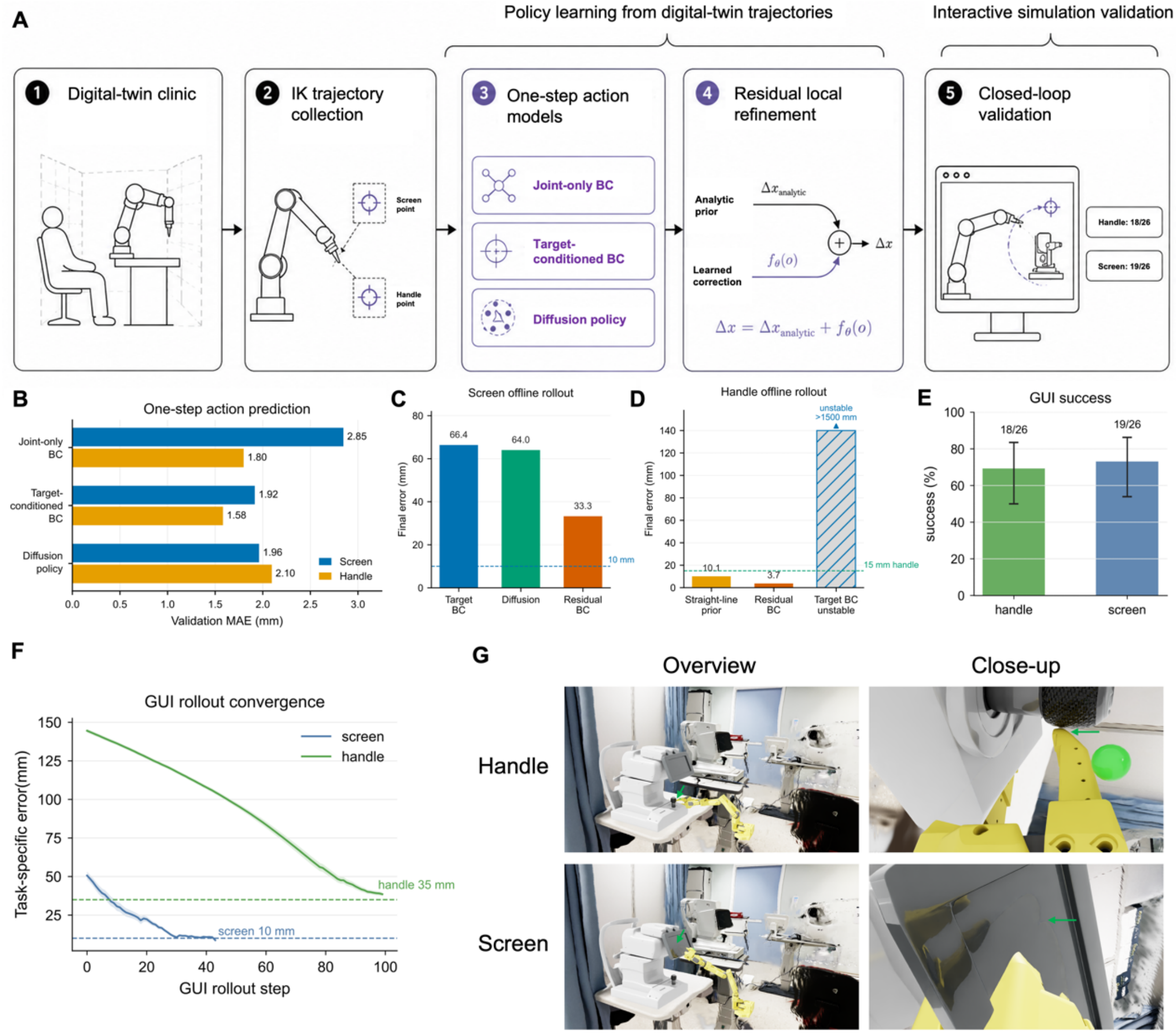

**Figure 6. Digital-twin trajectories support policy learning and closed-loop task evaluation.** (A) Workflow linking digital-twin construction, IK trajectory collection, policy learning, offline rollout and closed-loop simulation evaluation. Policy families included joint-only behaviour cloning, target-conditioned behaviour cloning, conditional diffusion policy and residual behaviour cloning. (B) One-step validation mean absolute error for screen and handle action prediction. (C,D) Offline rollout performance for screen and handle tasks using stored validation states and task-specific metrics. (E) Closed-loop rollout success for screen and handle tasks under task-specific success gates. (F) Reduction in task-specific error during closed-loop rollout. (G) Representative successful screen and handle rollouts. Green arrows indicate successful contact sites.

## Discussion

This study addresses a practical challenge in medical EAI: how to evaluate systems in realistic outpatient environments before they operate near patients, staff and clinical equipment. We show that single-image reconstruction can support more than visual scene representation. When combined with simulator conversion, mesh grounding and task-specific anchors, routine clinic images can become operational digital twins for trajectory collection, contact-proxy testing and policy evaluation. These findings shift the focus of clinical digital twins from visual realism alone to task-centred operational validity.

Most clinical simulators were designed for human training, procedural rehearsal or visual demonstration. Virtual reality platforms such as the Eyesi Surgical simulator have improved surgical performance and reduced complication rates in cataract surgery.[28,29] Similar benefits have also been reported for simulation platforms used in other surgical domains.[30-32] Simulation has also expanded to diagnostic training, including slit lamp examination and fundoscopy.[16,33] These systems are typically designed around standardized training tasks and purpose-built environments. Predeployment evaluation of medical EAI raises a different question: whether a specific robot-task system can operate within the spatial and contact constraints of a particular clinic. Our framework links site-specific scene reconstruction with simulator conversion, explicit device geometry and task anchors, providing a practical route from clinic images to robot-facing evaluation environments. This layered construction defines operational validity within a task-specific interaction envelope, with fidelity assessed in relation to the approach space, support surfaces, device relationships and contact regions required by the task.

Ophthalmology and optometry clinics provide a stringent testbed for this problem. Across the matched operational scenes, the same contact anchors produced different feasibility margins across robot embodiments; in the perturbation scene, controlled changes in instrument pose shifted some screen-contact trajectories across success boundaries, while handle contact remained comparatively stable. These patterns are consistent with prior reachability and capability-map studies showing that manipulation feasibility depends jointly on robot kinematics, base placement and the task-specific pose constraints of the target workspace.[34] Operational digital twins can therefore expose configuration-sensitive boundaries before physical testing and inform robot placement, task allocation or environment adjustment.

Single-image 3DGS served as the site-specific scene substrate within this layered construction.[27,35] It recovered the observed room appearance and spatial context from a single clinic image, preserving the visual and relative spatial relationships visible within the observed workspace. Explicit interaction geometry was concentrated around task-relevant regions through device meshes, colliders, articulation structures and semantic task anchors. This hybrid representation assigns distinct roles to different forms of fidelity: the Gaussian scene retains the wider clinical context, while mesh grounding supports reachability, collision and contact assessment. The resulting architecture concentrates modelling effort on interaction-critical regions and provides a practical route from visually reconstructed clinics to operational simulation environments.

Editability adds a complementary capability by enabling matched variants of the same clinic context. This capability is relevant to outpatient examination rooms, where flexible layouts, repositionable equipment and user orientation can influence clinical workflow.[36] Controlled environmental variation has also been used in robotics simulation to address robustness to changes in visual conditions, object pose and physical disturbances.[37] In this study, local image-space editing modified target-device regions while preserving the surrounding room and non-target equipment, allowing alternative reconstructions to be generated from the same source image. This separation between stable context and task-specific elements supports controlled comparisons in which selected scene factors change while the wider environment remains approximately constant. Image editing and mesh perturbation operate at different scales: the former creates alternative contextual configurations, whereas the latter probes contact feasibility around a specified device pose. Together, these operations turn a reconstructed clinic into an experimental substrate for testing configuration sensitivity before physical setup.

Beyond scene construction, the broader value of an operational digital twin lies in supporting staged robot development and evaluation. Prior studies have shown that simulation-generated trajectories can support policy learning across varied task configurations,[38] and that reconstructed or purpose-built simulated environments can be used to refine policies and reproduce aspects of physical policy performance when relevant control and visual discrepancies are addressed.[37,39] Our findings extend this approach to a clinical environment and support staged robot-facing evaluation before physical testing: task-grounded IK and closed-loop rollout provide complementary assessments of geometric feasibility and local control, while non-patient hardware execution provides an initial check that selected commands remain executable through the physical control stack. Future work should establish predictive validity by testing whether simulation-derived contact margins and failure boundaries correspond to physical outcomes across sites, robot embodiments and ophthalmic devices.

Several limitations should be considered. The present evaluation covered a limited set of ophthalmic and optometric scenes, so generalizability across clinical sites remains to be assessed. The framework primarily captures the spatial and geometric properties of clinical environments, while temporal variation, human-robot interaction and force-dependent device operation remain outside its current scope. The downstream experiments were designed as staged task-based evaluations under predefined conditions; the findings therefore establish operational feasibility within these conditions but do not determine autonomous performance, safety or clinical utility in routine care.

In conclusion, we developed and evaluated a clinic-derived operational digital-twin framework that converts routine ophthalmic images into editable, simulation-ready environments for task-specific evaluation of medical EAI. The staged workflow links site-specific reconstruction with contact-feasibility probing, controlled configuration changes, trajectory generation, policy evaluation and initial physical-robot testing, extending clinical scene reconstruction from visual representation to robot-relevant experimentation. Operational digital twins may therefore provide a practical intermediate layer between offline model development and prospective evaluation of embodied systems in clinical environments.

## Methods

### Study design

This study evaluated an image-derived workflow for constructing operational digital twins of outpatient ophthalmic clinics for simulation-based task evaluation of embodied robotic systems. The workflow linked single-image scene reconstruction, image-space editing, simulator conversion, clinical-device mesh integration, contact-geometry diagnostics, multi-arm contact-feasibility probing, controlled instrument-pose perturbation and local contact-policy evaluation. Analyses were organized by experimental layer, with scene inclusion narrowed according to the technical requirements of each downstream experiment. Reconstruction, depth and conversion analyses used the full 39-scene clinic cohort. Image-space editing was limited to the 32 scenes containing target ophthalmic devices. Mesh-integration analyses used seven scenes with collider-ready instrument assets and stable simulator placement. Multi-arm IK analyses used three matched operational scene variants with comparable robot placement and task anchors. Policy-learning and rollout analyses used the SO101 policy-evaluation scene, which had sufficient trajectory data for training and closed-loop testing. Layer-specific endpoints included observed-view and cross-view reconstruction agreement, prespecified conversion pass rates, contact-manifold readouts, task-specific IK contact-proxy success, strict and proxy perturbation outcomes, and task-specific closed-loop rollout success gates. The robot-facing analyses were therefore framed as simulation diagnostics for contact-proxy feasibility and local policy evaluation, rather than as tests of full autonomous manipulation, force-controlled device operation or real-world transfer. Selected simulator-derived trajectories and the handle-contact policy were subsequently evaluated on a non-patient SO101 platform through repeated physical replay and online inference execution, providing a bounded hardware check of the entire digital-twin clinic workflow.

### Clinical scene dataset

The dataset comprised 39 RGB routine photographs collected from real-world outpatient ophthalmic clinic environments, including examination rooms, waiting areas, treatment bays, and more open clinic layouts. Each scene was assigned a shared identifier linking the source image to evaluation-specific assets, including LiDAR depth data, capture-time depth visualizations, edited images, reconstruction outputs, simulator renderings, operational screenshots and trajectory records. Scene pairings were fixed before final aggregation so that reconstruction, conversion, editing and operational readouts could be traced to the same clinic-scene index.

### Single-image scene reconstruction

Each source photograph was reconstructed with a single-image 3DGS pipeline[27] to generate an editable scene representation. The reconstruction was treated as an operational scene representation of the photographed clinic environment rather than as a complete architectural model. For each scene, the reconstructed representation was exported for observed-view rendering, ordered orbit rendering and downstream simulator conversion. These outputs provided the common visual and geometric basis for the subsequent fidelity, conversion, editing and robot-facing analyses.

**Image-space scene editing**

When local reconfiguration was required, instrument-removal masks were generated for scenes containing ophthalmic devices using image-guided segmentation and AI-assisted, scene-specific mask generation.[40] Inpainting was then used to remove the instrument and approximate the local background before reconstruction was rerun. The edited image was processed with the same reconstruction pipeline as the unedited source image, producing paired raw and edited reconstructions under a matched workflow. Editing was assessed as a localized scene intervention: target-region change was separated from preservation of the surrounding scene, and paired raw-versus-edited outputs were retained for downstream preservation and stability analyses.

**Simulator conversion and scene instantiation**

Reconstructed scenes were converted into simulator-compatible assets and instantiated in Isaac Sim (v5.0; NVIDIA, USA).[41] Conversion produced simulator renderings and scene files for downstream inspection, cross-format comparison and robot-facing evaluation. For operational scenes, imported clinical-device meshes were scaled and aligned to the reconstructed room context, then assigned conservative collision proxies, pivots and semantic task anchors. These instantiated scenes provided the basis for mesh-integration analysis, contact-geometry diagnostics, IK probing, perturbation assays and closed-loop policy rollout. The full rendering, descriptor, and alignment workflow is described in the **Supplementary Methods**.

**Scene-level fidelity and geometric assessment**

Observed-view reconstruction quality was summarized with SSIM,[42] LPIPS,[43] CLIP similarity,[44] and PSNR.[45] DISTS[46] and FID[47] were reported as supportive perceptual and distributional descriptors. Higher SSIM, CLIP similarity, and PSNR indicate stronger agreement between the source image and the observed-view reconstruction. Lower LPIPS, DISTS, and FID indicate less perceptual or distributional drift. These metrics were interpreted jointly because they emphasize different failure modes, and none was used as a stand-alone acceptance gate.

Geometry was assessed using dominant-plane residuals, surface-normal error, and LiDAR-aligned depth metrics. Dominant-plane residuals were computed by fitting the largest support plane in each scene and summarizing the root-mean-square point-to-plane distance of assigned surface points; lower values indicate smoother large support surfaces and stronger room-scale geometric regularity. Surface-normal error measured the angular deviation between reconstructed dominant-surface normals and the fitted reference orientation, with lower values indicating less orientation drift in the reconstructed room envelope. LiDAR-aligned depth evaluation was performed after harmonizing orientation and scale between capture-time LiDAR depth arrays and rendered 3DGS depth maps. AbsRel, RMSE, MAE, and Log10 depth error summarized depth disagreement on overlapping valid pixels, with lower values indicating stronger depth agreement. The delta_1.25 metric summarized the fraction of valid pixels within the standard relative-error tolerance band, with higher values indicating more spatially consistent depth recovery. Alignment and aggregation details are provided in the **Supplementary Methods**.

**Multi-view coherence assessment**

Cross-view stability was evaluated from ordered circular render trajectories generated around each reconstructed scene. Consecutive views were sampled from the same orbit and camera-angle progression, so frame-to-frame differences reflected viewpoint change within one trajectory rather than mismatched views across scenes. Adjacent-view SSIM, PSNR, LPIPS and DISTS were computed along each trajectory to summarize structural and perceptual stability under camera motion. Higher adjacent-view SSIM and PSNR indicate stronger structural coherence, whereas lower adjacent-view LPIPS and DISTS indicate stronger perceptual consistency. Scene-level summaries were computed from the ordered frame sequences sampled from each rendered orbit.

**Conversion consistency assessment**

Simulator-conversion fidelity was summarized using three prespecified descriptive scores. Symmetric Chamfer similarity measured whether frames in one trajectory had close counterparts in the other trajectory after descriptor matching; after normalization, higher values indicate stronger cross-format visual correspondence.[48] Cyclic alignment similarity measured whether two ordered trajectories could be aligned by circular shift and optional reversal, with higher values indicating stronger sequence-level agreement. Structural preservation was summarized as 1 - smoothness gap, where the smoothness gap quantified the difference in average frame-to-frame descriptor displacement between source and converted sequences. Values closer to 1 indicate that the converted scene changes smoothly at approximately the same rate as the source sequence.

Per-scene conversion scores were first retained as continuous readouts and then binarized with prespecified descriptive cutoffs: $\geq 0.72$ for visual similarity, $\geq 0.65$ for cyclic alignment and $\geq 0.85$ for structural preservation. The structural-preservation cutoff corresponds to a smoothness-gap boundary of $\leq 0.15$. These cutoffs were fixed before final aggregation and were used for cohort stratification rather than post hoc threshold discovery.

**Editing preservation assessment**

Editing preservation was evaluated by separating intended target-region change from surrounding-scene preservation. Metrics were calculated for scenes with a visible target clinical instrument, a valid instrument mask and an analyzable paired edited output. The masked-region metric was interpreted as a descriptor of edited-area recovery and residual artifacts, whereas the unmasked-region metric reflected preservation of the surrounding scene. Scenes without a clearly defined instrument-removal target were excluded from this layer-specific analysis. Masked-region and unmasked-region SSIM quantified local edit effects and non-target context preservation, respectively. Global and local FID summarized whole-image and edited-region distribution shift. Raw and edited reconstructions were also compared using a composite scene-quality score derived from trajectory-level stability metrics. The absolute raw-versus-edited score shift was used as a within-study preservation index and was computed as the absolute difference between standardized composite scores for paired raw and edited reconstructions.

**Clinical-device mesh integration**

Clinical-device meshes and static objects were imported into selected reconstructed rooms after simulator instantiation. Meshes were prepared separately, scaled and aligned to the reconstructed room context, then assigned simplified collision geometry, pivots and task anchors. Details of device-mesh preparation and collision-proxy configuration are provided in the **Supplementary Methods**. Collision geometry was intentionally conservative to support stable simulator interaction and contact-proxy queries rather than exact mesh-level contact physics. Visible collider occupancy was measured from simulator screenshots to assess whether debug-rendered physics proxies intruded into the user-visible scene. The 7% reference line used in display items was a practical visualization guide for contextualizing visible interference, not a universal acceptance threshold.

**Contact diagnostics**

Operational reachability trajectories were recorded in the simulator and filtered to identify low-speed contact windows. Dominant contact clusters were identified, and PCA was used to summarize the geometry of the dominant interaction manifold. Dispersion radius measured spread around the contact centroid, whereas PCA-derived in-plane and normal-direction spreads measured variation within and away from the dominant task plane. Plane-fit RMSE summarized residual distance to the fitted plane, and the normal-to-in-plane dispersion ratio provided a dimensionless measure of planar concentration. Lower dispersion radius, lower plane-fit RMSE and lower normal-to-in-plane ratio indicate tighter, more planar interaction geometry. These metrics were interpreted as contact-manifold diagnostics rather than force measurements.

**Multi-arm contact-feasibility probing**

Multi-arm IK probing tested whether mesh-integrated digital-twin clinic scenes could be queried by different robotic embodiments under shared task intent. The assay queried SO101, Kinova Gen3 and Franka Panda across matched operational scene variants. Robot-arm configurations are summarized in **Supplementary Table 6**. Each robot used its own kinematic adapter, active-joint set, joint limits, end-effector frame and base-placement convention. For each robot-scene-task cell, the solver sampled a local anchor neighborhood using target, seed and approach jitter. The primary denominator was the final benchmark complete episode, defined as an exported acquisition row with an associated episode file from the final benchmark source. Historical tuning, smoke, pilot and debug batches were excluded from the main denominator.

Task-specific contact-proxy success was the primary IK endpoint, while final positional error was retained only as a secondary anchor-deviation diagnostic. Screen probing used signed screen-plane residual and in-plane error, with candidate probes evaluated against configured contact margins where applicable. Handle probing used a signed bbox/probe mesh-proximity readout: positive values indicated residual separation outside the proxy volume, whereas negative values indicated probe entry. This readout should be interpreted as a geometric proximity proxy, not an exact triangular-mesh signed distance, contact-sensor signal or force-contact measurement. Failure categories were grouped after final episode aggregation. IK failures were interpreted as configuration-sensitive feasibility outcomes shaped by robot geometry, placement, target location, collision constraints and solver behavior.

**Controlled instrument-pose perturbation assay**

The perturbation assay tested configuration sensitivity in robot-specific instantiations of a representative clinic layout. Seven instrument-pose conditions were evaluated: baseline, local x -20 mm, local x +20 mm, local y -20 mm, local y +20 mm, yaw -5 degrees and yaw +5 degrees. Planar translations were applied in the instrument-local coordinate frame, and yaw rotations were applied around the instrument root local origin. The instrument was reset from the nominal pose before each independent condition, and screen and handle anchors were recomputed after each transform. Each robot-task-condition used 20 jittered trials.

Strict task-native success was defined by the task script overall-success field. Task-proxy success was reported separately to distinguish strict simulator task completion from contact-relevant geometric evidence. For screen contact, task-proxy success was the logical union of screen point success, screen mesh-touch success and screen center-touch success. For handle contact, the common task-proxy metric was contact-proxy success. Diagnostic readouts included endpoint error, solver iterations, failure reason, screen-plane residual, in-plane error and handle bbox/probe contact-proxy distances.

**Local contact-policy learning**

Policy learning was formulated as supervised imitation learning from SO101 digital-twin trajectories in the policy-evaluation scene. Benchmark IK trials and policy trajectories were treated as distinct data units: benchmark trials summarized contact-proxy feasibility at task anchors, whereas policy trajectories provided time-indexed state-action examples for local contact refinement. Actions were represented as next-step 3D end-effector displacement. Screen training used local pre-contact and contact frames from IK capture trajectories. Handle training used prefiltered handle-contact trajectories with fixed train-validation splits.

Four policy families were evaluated. Joint-only behavior cloning predicted end-effector displacement from joint state. Target-conditioned behavior cloning added the target-relative vector. Diffusion policy modeled the same low-dimensional action through conditional denoising. Residual behavior cloning decomposed the command into an analytic local controller plus a learned residual correction. These models were evaluated as local contact-policy components in simulation, not as general-purpose autonomous clinical manipulation systems.

**Offline and closed-loop rollout evaluation**

Evaluation proceeded in three stages: one-step supervised validation, offline rollout and closed-loop simulation rollout. One-step validation assessed supervised next-action prediction on held-out frames. Offline rollout iteratively applied predicted end-effector displacements to stored validation states without runtime simulator rendering or contact stepping. Closed-loop simulation rollout reset SO101 to validation initial states in the policy-evaluation scene, computed task targets from runtime scene geometry, performed CPU policy inference, applied end-effector deltas through finite-difference IK within the simulator loop, and recorded synchronized environment and wrist-camera images.

Task-specific rollout success gates were fixed before reporting. Closed-loop screen success used a 10 mm point-alignment gate to the screen target. Closed-loop handle success used the calibrated bbox-plus-probe contact gate, requiring bbox distance $\leq 5$ mm and probe distance $\leq 35$ mm. These gates were used to evaluate local simulated contact behavior in the digital-twin scene. Because the screen and handle gates were task-specific, they should not be interpreted as equivalent physical error metrics.

**Physical replay and online policy execution**

Physical testing used a non-patient SO101 platform. Replay was performed beside ophthalmic equipment in a clinic-like arrangement, whereas online policy execution used a simplified tabletop handle target. The planned cohort comprised three handle and three screen NPZ trajectories, each replayed five times, and five handle-policy executions. A single-view video and terminal log were recorded for each planned run.

For replay, the NPZ joint sequence was published and the robot subscriber converted the six commanded joint angles from radians to servo ticks before issuing target commands. Online execution subscribed to live encoder-derived joint signals. At each step, a runtime-loaded URDF supplied forward kinematics for the nominal end-effector position; the target-relative observation drove a Cartesian policy increment, and a numerical Jacobian produced candidate joint commands for publication on joint states. Logged nominal target distance was normalized within run as $d(t)/d0$, where $d0$ was the first valid logged value. Visible target approach and contact were assessed qualitatively from single-view video.

**Statistical analysis**

All summary statistics are reported as mean ± SEM unless otherwise stated. Scene-impact preservation after editing was summarized descriptively using absolute score differences and the proportion of scenes below the prespecified descriptive threshold. Scene conversion was analyzed using continuous scene-level metrics and thresholded pass rates because the goal was fidelity assessment rather than intervention-versus-control inference. Multi-arm IK, perturbation and policy-rollout analyses were reported as task-specific simulation diagnostics, with denominators defined before final aggregation and success rates reported with numerator and denominator counts where relevant. Statistical analyses were performed in Python.

**Acknowledgments**

We thank the InnoHK HKSAR Government for providing valuable support.
This work described in this paper was conducted in the JC STEM Lab of Innovative Light Therapy for Eye Diseases funded by The Hong Kong Jockey Club Charities Trust.

**Funding**

D.S. and M.H. disclose support for the review and publication of this work from the RCSV seed fund Eye Robot for Autonomous Clinic: Prototype Development (P0057912) from PolyU.

**Author Contributions**

X.W. designed and developed the study workflow, constructed the digital-twin clinic simulation environment, implemented the downstream validation pipelines, performed the experiments and analyses, prepared the figures, and wrote the initial manuscript. J.Z. provided algorithmic support and assisted with framework implementation. M.X. and H.K.C. provided technical guidance, project coordination, and manuscript review. M.H. and D.S. contributed to study conception and design, supervision, project oversight, and manuscript review and editing. All authors critically reviewed the manuscript, provided important intellectual input, and approved the final version.

**Competing Interests**

The authors declare no conflicts of interest.

**Data Availability**

The data supporting the findings of this study are available from the corresponding authors upon reasonable request. Source clinic images are subject to institutional and site-privacy restrictions.

**Code Availability**

Custom code used for the downstream robotic experiments is available from the corresponding authors upon reasonable request. The principal third-party software packages used in this study are publicly available: SHARP for single-image 3DGS reconstruction (https://github.com/apple/ml-sharp), 3DGRUT for Gaussian-scene conversion and rendering (https://github.com/nv-tlabs/3dgrut), SAM 2 for image segmentation (https://github.com/facebookresearch/sam2), and NVIDIA Isaac Sim 5.0 for robotic simulation (https://docs.isaacsim.omniverse.nvidia.com/).